\documentclass{article}
\usepackage{amsmath,graphicx,mlspconf}
\usepackage{amssymb}
\usepackage{booktabs}
\usepackage{xcolor}

\copyrightnotice{U.S.\ Government work not protected by U.S.\ copyright}

\copyrightnotice{979-8-3503-2411-2/25/\$31.00 {\copyright}2025 Crown}

\copyrightnotice{979-8-3503-2411-2/25/\$31.00 {\copyright}2025 European Union}

\copyrightnotice{979-8-3503-2411-2/25/\$31.00 {\copyright}2025 IEEE}

\toappear{2026 IEEE International Workshop on Machine Learning for Signal Processing, Sep.\ 28-- Oct.\ 1, 2026, Atlanta, USA}

\title{Disentangling Co-Occurring Retinal Pathologies with Saliency-Guided Sparse Expert Routing}

\name{Nagur Shareef Shaik$^{\star}$, Jeongwoo Park$^{\dagger}$, Yeong-Jin Kim$^{\dagger}$, Jaeuk Jung$^{\dagger}$, Hyunjung Oh$^{\dagger}$, Dong Hye Ye$^{\star}$}

\address{$^{\star}$ Department of Computer Science, Georgia State University, Atlanta, GA 30324, USA \\ $^{\dagger}$ Piehealthcare Inc., Suwon-si, Gyeonggi-do, 16311, Republic of Korea
}

\begin{document}

\maketitle

\begin{abstract}
Retinal fundus images frequently exhibit multiple co-occurring pathologies, yet standard deep learning classifiers apply static, identical computation to every image regardless of the underlying disease distribution. We propose a novel architecture that resolves this via sparse conditional computation, pairing a Guided Context Gating (GCG) spatial attention front-end with a sparsely-routed Mixture-of-Experts (MoE) block operating over feature tokens. Crucially, this routing yields an interpretable, data-driven decomposition. Expert allocation is significantly disease-dependent ($p < 0.001$), with the healthy ``Normal'' state and morphologically distinct pathologies (e.g., ERM, AMD) isolating to dedicated experts. On a five-class, patient-disjoint 5-fold cross-validation benchmark, our model achieves $0.912 \pm 0.008$ macro AUC and $0.653 \pm 0.014$ macro F1. Furthermore, Grad-CAM++ and post-MoE t-SNE visualizations confirm that expert routing aligns with localized lesions and geometrically maps co-occurring cases between their constituent clusters, positioning sparse MoE as an interpretable approach to multi-disease retinal screening.
\end{abstract}

\newcommand{\cem}[1]{\textcolor{blue}{cem: #1}}
\section{Introduction}
\label{sec:intro}

Color fundus photography is the primary non-invasive modality for screening vision-threatening conditions such as diabetic retinopathy (DR), age-related macular degeneration (AMD), epiretinal membrane (ERM), and glaucoma \cite{li2019diagnostic, bhati2023discriminative}. In practice, these conditions are not mutually exclusive: particularly in aging and diabetic populations, a single eye may present several pathologies at once, for instance, the microaneurysms of DR alongside the optic-disc cupping of glaucoma. This makes \emph{multi-label} classification the clinically faithful, but considerably harder formulation, relative to the single-disease setting that dominates the literature \cite{he2016deep, dosovitskiy2020image, shaik2025ordinal}.

Most deep classifiers rely on a single, shared feature extractor, whether through standard transfer-learned ensembles \cite{gour2021multi, wang2020multilabel} or architectures explicitly designed to model label correlations \cite{li2020dense, bhati2023discriminative}. This forces the network to encode distinct, and sometimes conflicting, disease morphologies inside one dense embedding. Consequently, when pathologies co-occur, their representations become \emph{entangled}, degrading both accuracy and interpretability. Attention mechanisms, such as channel and spatial gating \cite{hu2018squeeze, woo2018cbam}, non-local context \cite{wang2018non, cao2019gcnet}, or lesion-guided gating \cite{cherukuri2024guided}, help by emphasizing salient regions, but they do not change a more basic property of the computation: it is \emph{static}. Every spatial location is processed by the same feed-forward weights, whether it contains a hemorrhage, drusen, or healthy retina. The model's \emph{what} to attend to is adaptive; its \emph{how} to compute is not.

Sparse mixture-of-experts (MoE) layers relax exactly this constraint through \emph{conditional computation}: a lightweight router dispatches each token to a small subset of expert sub-networks, allowing capacity to grow without a proportional growth in per-image computational cost \cite{shazeer2017outrageously, fedus2022switch}. MoE is now standard in large language models, yet in vision, it has been used mainly as a scaling device for single-label, natural-image classification \cite{riquelme2021scaling}. Its potential as a mechanism for \emph{disentangling} co-occurring medical pathologies, and the question of whether expert routing aligns with \emph{disease identity}, remains largely unexplored.

We propose a compact pipeline that shifts fundus analysis from static processing toward pathology-aware routing. A Guided Context Gating (GCG) front-end \cite{cherukuri2024guided} first reweights the backbone feature map toward lesion-salient regions, ensuring that tokenization feeds the router clinically relevant evidence rather than background anatomy. The resulting spatial tokens are processed by a single sparsely-routed MoE transformer block that adapts the modern language-model recipe, top-$k$ routing, SwiGLU experts, and RMSNorm \cite{shazeer2020glu, zhang2019root}, to the medical setting. Following DeepSeekMoE \cite{dai2024deepseekmoe}, an always-on \emph{shared} expert absorbs common retinal structures (e.g., optic disc, vasculature), freeing the routed experts to specialize on the residual variation that distinguishes diseases. Crucially, we do not treat the router as a black box: we quantify per-expert load conditioned on disease labels and relate it to classification behavior, turning the routing distribution into an interpretable diagnostic signal. 


\begin{figure*}[t]
  \centering
  \includegraphics[width=\textwidth]{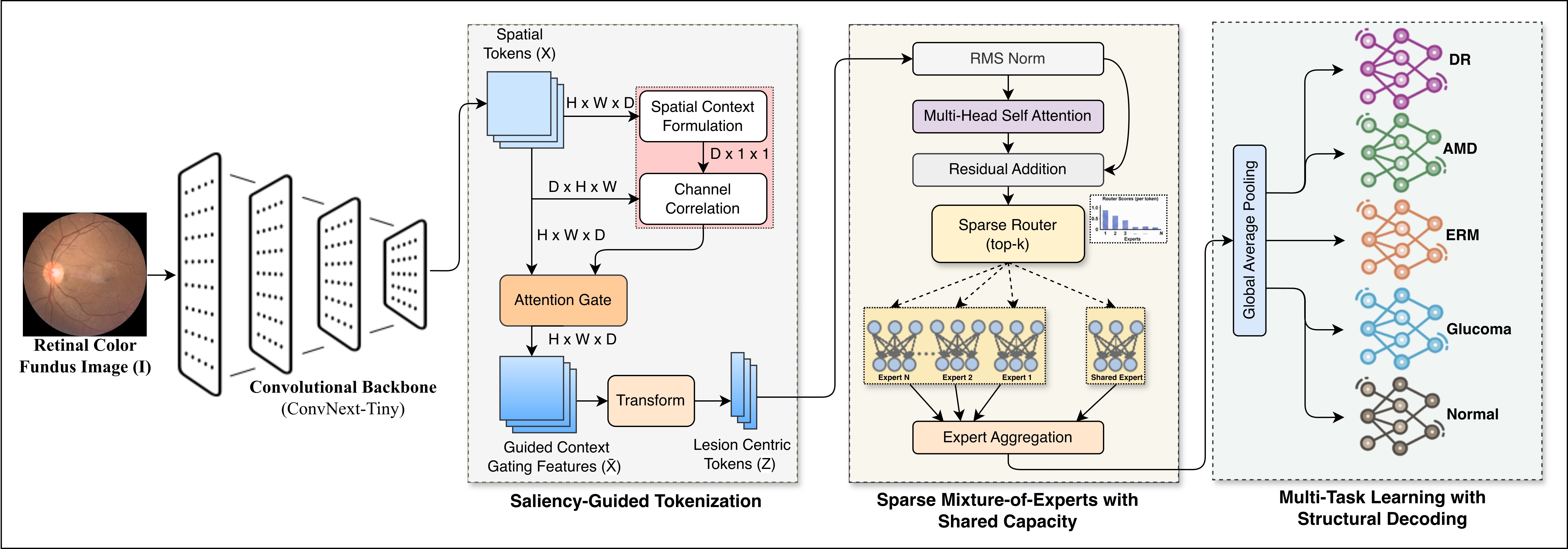} 
  \caption{Overview of the proposed Saliency-Guided Sparse Expert Routing architecture. A ConvNeXt-Tiny backbone extracts dense spatial features. \textbf{(1) Saliency-Guided Tokenization:} The Guided Context Gating (GCG) module recalibrates features based on global lesion saliency before flattening them into tokens. \textbf{(2) Sparse MoE with Shared Capacity:} A router dynamically dispatches tokens to the top-2 specialized experts, while a DeepSeek-style Shared Expert processes all tokens to anchor common retinal anatomy. \textbf{(3) MTL with Structural Decoding:} Following Global Average Pooling (GAP), independent Multi-Task Learning heads generate class logits, which are projected onto a clinically valid manifold via Ontological Decoding.}
  \label{fig:architecture}
\end{figure*}

\section{Methodology}
\label{sec:method}

We formulate multi-disease retinal screening as a multi-label classification task. Let $\mathcal{D} = \{(\mathbf{I}^{(i)}, \mathbf{y}^{(i)})\}_{i=1}^N$ denote a dataset where $\mathbf{I} \in \mathbb{R}^{3 \times H_0 \times W_0}$ is a high-resolution fundus image and $\mathbf{y} \in \{0, 1\}^C$ is the corresponding multi-hot label vector for $C=5$ clinical states (DR, AMD, ERM, Glaucoma, Normal). Our objective is to learn a mapping $f_\theta: \mathbf{I} \rightarrow [0, 1]^C$ that reduces the feature entanglement induced by co-occurring pathologies through pathology-aware conditional computation.
First, a pre-trained convolutional backbone (ConvNeXt-Tiny) projects the input $\mathbf{I}$ onto a dense spatial manifold, yielding a feature tensor $\mathbf{X} \in \mathbb{R}^{B \times D \times H \times W}$, where $B$ is the batch size, $D$ is the channel dimension, and $H, W$ are the spatial dimensions. To optimally adapt this manifold to the fundus domain while preserving low-level Gabor-like filters, we freeze the initial 80\% of the backbone parameters and fine-tune the terminal 20\%.

\subsection{Saliency-Guided Tokenization}
\label{ssec:gcg}
Uniform tokenization in standard Vision Transformers blindly flattens spatial patches, forcing subsequent routing mechanisms to expend equal computational bandwidth on spurious background anatomy and minute lesions. We employ Guided Context Gating (GCG) \cite{cherukuri2024guided}, that acts as a spatial information bottleneck, recalibrating $\mathbf{X}$ based on global lesion saliency prior to sequence generation. The GCG module operates via two sequential mechanisms: Global Context Attention and Attention Gating.

\subsubsection{Global Context Attention (GCA)}
The GCA module synthesizes a global context descriptor by aggregating spatial features weighted by their pathological relevance. Let $\mathbf{x}_{i,j} \in \mathbb{R}^D$ denote the feature vector at spatial location $(i, j)$. A $1 \times 1$ convolution parameterized by $\mathbf{W}_c \in \mathbb{R}^{1 \times D}$ computes a raw spatial attention map, which is normalized via a spatial softmax operator:
\begin{equation}
    m_{i,j} = \frac{\exp(\mathbf{W}_c \mathbf{x}_{i,j})}{\sum_{u=1}^H \sum_{v=1}^W \exp(\mathbf{W}_c \mathbf{x}_{u,v})}
\end{equation}
The global context vector $\mathbf{g} \in \mathbb{R}^D$ is then obtained via weighted global pooling: $\mathbf{g} = \sum_{i=1}^H \sum_{j=1}^W m_{i,j} \mathbf{x}_{i,j}$. 

To capture complex cross-channel dependencies, $\mathbf{g}$ undergoes a non-linear bottleneck transformation. Let $\mathbf{W}_{t1} \in \mathbb{R}^{\frac{D}{r} \times D}$ and $\mathbf{W}_{t2} \in \mathbb{R}^{D \times \frac{D}{r}}$ be projection matrices with a reduction ratio $r=8$. The transformed context is:
\begin{equation}
    \mathbf{g}' = \mathbf{W}_{t2} \left( \text{LN} \left( \max(0, \mathbf{W}_{t1} \mathbf{g}) \right) \right)
\end{equation}
where $\text{LN}(\cdot)$ denotes Layer Normalization. The context-enriched feature map is formed via a spatial broadcast addition: $\mathbf{X}_{ctx} = \mathbf{X} \oplus \mathbf{g}'$, where $\oplus$ denotes addition with broadcasting across the spatial dimensions $H$ and $W$.

\subsubsection{Attention Gate (AG)}
The AG fuses the local features $\mathbf{X}$ and the global context $\mathbf{X}_{ctx}$ to generate a highly selective spatial gating map $\mathbf{A} \in [0, 1]^{B \times 1 \times H \times W}$. Let $\circledast$ denote the convolution operator. The intermediate fusion tensor $\mathbf{F} \in \mathbb{R}^{B \times D \times H \times W}$ is computed as:
\begin{equation}
    \mathbf{F} = \max(0, \mathbf{W}_x \circledast \mathbf{X} + \mathbf{W}_g \circledast \mathbf{X}_{ctx} + \mathbf{b}_{xg})
\end{equation}
where $\mathbf{W}_x, \mathbf{W}_g \in \mathbb{R}^{D \times D \times 1 \times 1}$ are learnable kernels and $\mathbf{b}_{xg} \in \mathbb{R}^D$ is a bias vector. The spatial attention map is then derived by projecting the channel-normalized fusion tensor:
\begin{equation}
    \mathbf{A} = \sigma \left( \mathbf{W}_\psi \circledast \text{LN}_{c}(\mathbf{F}) + b_\psi \right)
\end{equation}
where $\mathbf{W}_\psi \in \mathbb{R}^{1 \times D \times 1 \times 1}$, $b_\psi \in \mathbb{R}$, $\sigma(\cdot)$ is the logistic sigmoid function, and $\text{LN}_c$ denotes Layer Normalization applied strictly across the channel dimension. 

The gated features $\tilde{\mathbf{X}} = \mathbf{A} \odot \mathbf{X}$ (where $\odot$ is the Hadamard product) are then flattened along the spatial dimensions and transposed to form a sequence of lesion-centric tokens $\mathbf{Z} \in \mathbb{R}^{B \times S \times D}$, where the sequence length is $S = H \times W$.

\subsection{Sparse Mixture-of-Experts with Shared Capacity}
\label{ssec:moe}
The token sequence $\mathbf{Z}$ is ingested by the Image MoE Block. To guarantee training stability under sparse routing dynamics, we employ Root Mean Square Normalization (RMSNorm) \cite{zhang2019root} in a pre-norm configuration. For a vector $\mathbf{x} \in \mathbb{R}^D$, RMSNorm is defined as:
\begin{equation}
    \text{RMSNorm}(\mathbf{x}) = \frac{\mathbf{x}}{\sqrt{\frac{1}{D}\sum_{i=1}^D x_i^2 + \epsilon}} \odot \boldsymbol{\gamma}
\end{equation}
where $\boldsymbol{\gamma} \in \mathbb{R}^D$ is a learnable scale parameter and $\epsilon = 10^{-6}$. The sequence first traverses a Multi-Head Self-Attention (MHSA) layer to model long-range spatial dependencies: $\mathbf{Z}' = \mathbf{Z} + \text{MHSA}(\text{RMSNorm}(\mathbf{Z}))$.

\textbf{Sparse Routing Mechanism:} Let $\mathbf{z}_t \in \mathbb{R}^D$ denote the $t$-th token in $\mathbf{Z}'$. A linear router parameterized by $\mathbf{W}_r \in \mathbb{R}^{D \times E}$ computes routing logits $\mathbf{h}_t = \mathbf{z}_t \mathbf{W}_r$ across $E=8$ independent experts. We enforce a strict top-$k$ routing policy ($k=2$). The set of selected expert indices for token $t$ is defined as:
\begin{equation}
    \mathcal{I}_t = \mathop{\mathrm{arg\,top}}_{i \in \{1, \dots, E\}}^k (h_{t,i})
\end{equation}
The routing weights are derived via a sparse softmax distribution over the selected subset:
\begin{equation}
    w_{t, i} = \frac{\exp(h_{t,i})}{\sum_{j \in \mathcal{I}_t} \exp(h_{t,j})}, \quad \forall i \in \mathcal{I}_t
\end{equation}

\textbf{Representation Disentanglement via Shared Capacity:} A known difficulty in MoE models is uneven expert utilization, where a few experts dominate while others model ubiquitous background structure. To mitigate this, we integrate a shared expert $E_{shared}$ \cite{dai2024deepseekmoe} that processes every token, acting as a structural anchor for common retinal anatomy and encouraging the routed experts to capture the \textit{residual pathological variation} that distinguishes diseases. This yields experts selectively associated with specific pathologies, though some capacity remains under-utilized in the absence of an explicit load-balancing objective. The final token representation is the sum of the shared baseline and the capacity-weighted routed experts:
\begin{equation}
    \mathbf{z}''_t = \mathbf{z}'_t + E_{shared}(\mathbf{z}'_t) + \sum_{i \in \mathcal{I}_t} w_{t, i} E_i(\mathbf{z}'_t)
\end{equation}

Each expert $E(\cdot)$ is parameterized as a SwiGLU feed-forward network \cite{shazeer2020glu}. Let $\text{SiLU}(x) = x \cdot \sigma(x)$. For an input $\mathbf{x}$, the expert computes:
\begin{equation}
    E(\mathbf{x}) = \text{Dropout}\left( \left( \text{SiLU}(\mathbf{x}\mathbf{W}_1) \odot (\mathbf{x}\mathbf{W}_3) \right) \mathbf{W}_2 \right)
\end{equation}
where $\mathbf{W}_1, \mathbf{W}_3 \in \mathbb{R}^{D \times 4D}$ project the input to a higher-dimensional space, and $\mathbf{W}_2 \in \mathbb{R}^{4D \times D}$ projects it back.


\subsection{Multi-Task Learning with Structural Decoding}
\label{ssec:mtl_heuristics}
Following the MoE block, Global Average Pooling (GAP) collapses the sequence dimension to yield a holistic image-level embedding $\mathbf{v} = \frac{1}{S} \sum_{t=1}^S \mathbf{z}''_t \in \mathbb{R}^D$. Because co-occurring pathologies require highly non-linear, intersecting decision boundaries, we project $\mathbf{v}$ through a decoupled Multi-Task Learning (MTL) head. The embedding is routed to $C$ independent Multi-Layer Perceptrons (MLPs), each featuring a dimensionality reduction cascade ($\mathbb{R}^D \to \mathbb{R}^{512} \to \mathbb{R}^{256} \to \mathbb{R}^1$) interleaved with 1D Batch Normalization, ReLU activations, and aggressive Dropout ($p \in \{0.3, 0.4\}$).

The network is optimized via a dynamically weighted Binary Cross-Entropy (BCE) loss to counteract extreme clinical class imbalance. The positive weight $w_c$ for class $c$ is computed as:
\begin{equation}
    w_c = \text{clip}\left( \sqrt{\frac{N_{\text{total}} - N_c}{N_c + \epsilon}}, \; 1.0, \; 5.0 \right)
\end{equation}
where $N_{\text{total}}$ is the total number of training samples and $N_c$ is the frequency of class $c$.

\textbf{Ontological Decoding:} 
Standard BCE assumes conditional label independence, ignoring the clinical mutual exclusivity between healthy and pathological states. To bridge this, we project the continuous probabilities $\hat{\mathbf{y}} = \sigma(\text{logits})$ onto a clinically valid boolean manifold $\mathbf{b} \in \{0, 1\}^C$ via structured decoding. Let $\mathcal{P} = \{1, 2, 3, 4\}$ denote pathological indices and $h = 5$ the healthy state. We enforce two strict priors:
\begin{enumerate}
    \item \textit{Ontological Exclusivity:} Any pathology strictly negates the healthy state, enforced via $b_h = \mathbb{I}(\hat{y}_h > 0.5) \cdot \mathbb{I}(\max_{c \in \mathcal{P}} \hat{y}_c \le 0.5)$.
    \item \textit{Maximum Likelihood Resolution:} Under extreme uncertainty ($\max(\hat{\mathbf{y}}) \le 0.5$), we force a prediction for the most probable class: $b_{c^*} = 1$ where $c^* = \text{argmax}(\hat{\mathbf{y}})$.
\end{enumerate}
Decoupling these hard constraints from the loss function allows the MoE block to learn continuous feature entanglements without being penalized by logical contradictions during training.

\section{Experiments \& Results}
\label{sec:results}

\subsection{Experimental Setup}
\label{ssec:setup}

\noindent\textbf{Dataset.}
We derive a five-class multi-label benchmark from the ODIR 2019 grand challenge dataset \cite{li2021benchmark}, focusing on conditions that frequently co-occur: $\{$ DR, AMD, ERM, Glaucoma, Normal$\}$. We exclude the remaining ODIR categories due to severe under-representation that destabilizes training. Left and right eyes are treated independently, and instances lacking target labels are discarded. Because this label space departs from the eight-class ODIR taxonomy, we re-train representative published leaderboard methods under our identical protocol for fair comparison. To ensure robust evaluation and prevent left/right-eye data leakage, all experiments utilize a strict \emph{patient-disjoint} 5-fold cross-validation protocol. Each cross-validation test fold contains $\sim$1{,}185 label instances with pronounced imbalance (Glaucoma $n{=}54$, AMD $n{=}56$ vs.\ Normal $n{=}583$; a $\sim$10.8$\times$ ratio).

\noindent\textbf{Implementation Details.}
The backbone is ConvNeXt-Tiny \cite{liu2022convnet} at $518\times518$ resolution, with the first $80\%$ of parameters frozen. The MoE block utilizes $E=8$ experts, top-$k=2$ routing, a shared expert, $8$ attention heads, and RMSNorm ($\epsilon=10^{-6}$). We optimize a class-weighted BCE for $100$ epochs per fold using AdamW (head lr $5\!\times\!10^{-5}$, backbone lr $5\!\times\!10^{-6}$, weight decay $10^{-2}$), cosine annealing, batch size $8$, and mixed precision. Metrics use a fixed per-class threshold of $0.5$. The model contains $97.85$M parameters, requires $58.15$ GFLOPs/sample, and runs at $5.23$\,ms/image on a single NVIDIA RTX 3090 GPU.

\subsection{Quantitative Results}
\label{ssec:quant}

\begin{table}[t]
\centering
\caption{Comparison with state-of-the-art methods with proposed approach emphasizing the effectiveness of disentangling co-occurring retinal pathologies (mean$\pm$std).}
\label{tab:sota}
\small
\setlength{\tabcolsep}{7pt}
\begin{tabular}{lcc}
\toprule
Method & mAUC & mF1 \\
\midrule
VGG16 Baseline \cite{li2021benchmark}      & $0.852\pm0.011$ & $0.545\pm0.016$ \\
TL-CNN \cite{gour2021multi}                & $0.868\pm0.010$ & $0.572\pm0.015$ \\
EfficientNet Ensemble \cite{wang2020multilabel} & $0.875\pm0.009$ & $0.586\pm0.014$ \\
DCNet \cite{li2020dense}                   & $0.882\pm0.010$ & $0.594\pm0.015$ \\
DKCNet \cite{bhati2023discriminative}      & $0.895\pm0.008$ & $0.615\pm0.013$ \\
\midrule
\textbf{Ours}                   & \textbf{0.912$\pm$0.008} & \textbf{0.653$\pm$0.014} \\
\bottomrule
\end{tabular}
\end{table}

We benchmark our architecture against representative published methods from the ODIR-2019 grand challenge. The compared methods span the dominant design families on this benchmark. As shown, the standard VGG16 Baseline \cite{li2021benchmark}, transfer-learned CNNs (TL-CNN) \cite{gour2021multi}, and EfficientNet Ensembles \cite{wang2020multilabel} plateau around $0.87$ AUC and $0.58$ F1, struggling with feature entanglement. Architectures explicitly designed to model label correlations, such as DCNet \cite{li2020dense} and the discriminative-kernel network DKCNet \cite{bhati2023discriminative}, perform better, reaching up to $0.895$ AUC. However, our proposed architecture outperforms all baselines ($+0.017$ mAUC and $+0.038$ mF1 over DKCNet; \textbf{see significance tests above}), indicating that dynamic conditional computation is better suited than static dense networks for disentangling co-occurring pathologies.

\begin{figure}[t]
  \centering
  \includegraphics[width=0.49\linewidth]{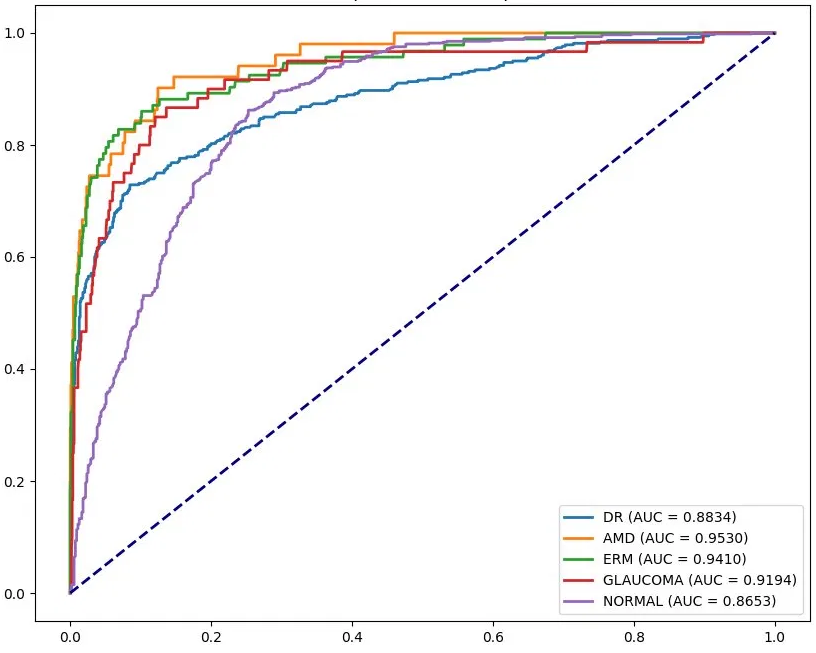}\hfill
  \includegraphics[width=0.49\linewidth]{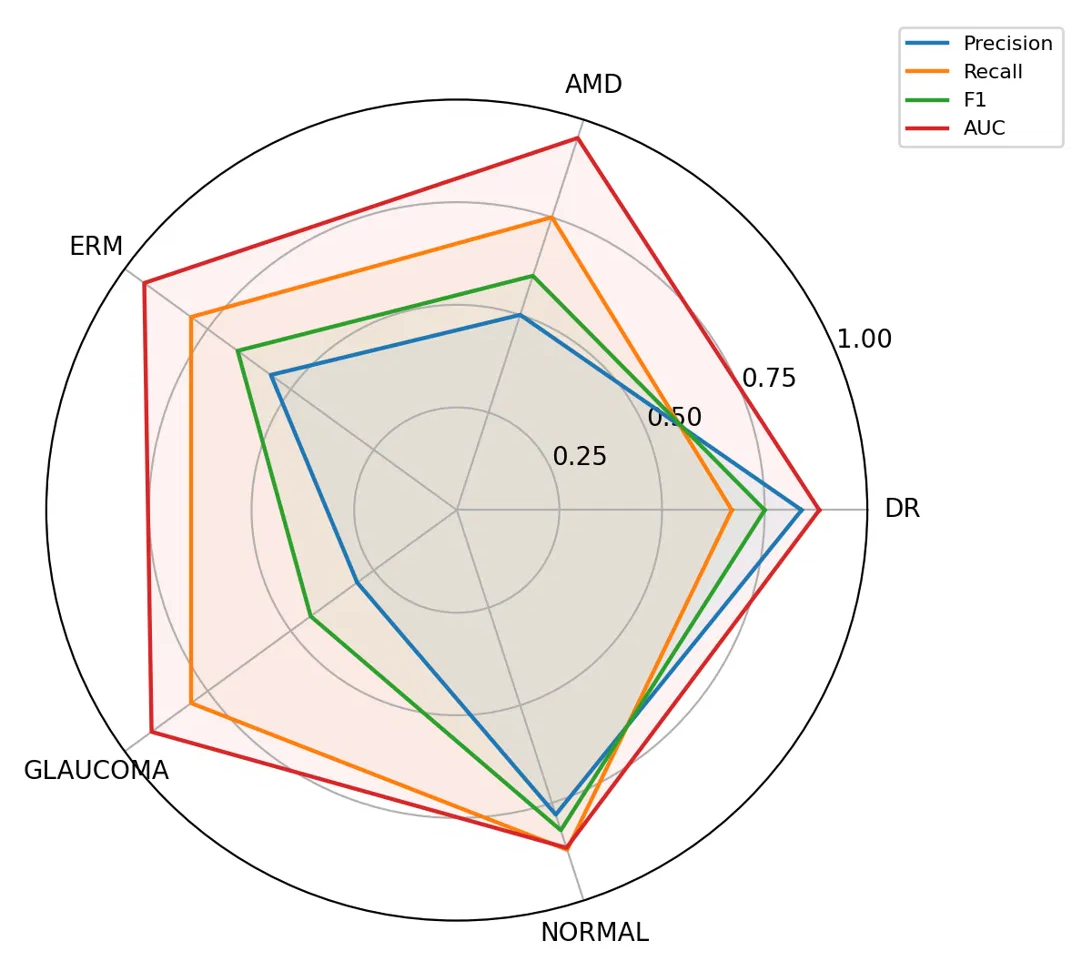}
  \caption{Quantitative performance. \textbf{Left:} Per-class ROC curves, showing high AUC for rare classes (AMD, ERM, Glaucoma). \textbf{Right:} Radar plot of per-class metrics; the AUC contour dominates the precision contour, illustrating a strong ranking capability despite threshold calibration gaps.}
  \label{fig:metrics}
\end{figure}

Across five folds, our model achieves a macro AUC of $0.912\pm0.008$, macro F1 of $0.653\pm0.014$, and accuracy of $0.649\pm0.011$. The sensitivity-leaning operating point (macro recall $0.776$ vs.\ precision $0.597$) is highly appropriate for clinical screening. Fig.~\ref{fig:metrics} details per-class performance. Notably, rare classes (AMD, ERM, Glaucoma) exhibit modest precision at the $0.5$ threshold but exceptionally high AUCs ($0.953$, $0.941$, $0.919$). As illustrated by the diverging radar contours, the model \emph{ranks} these pathologies highly; the lower precision reflects an uncalibrated global threshold for low-prevalence classes rather than a representational failure.

\begin{table}[t]
\centering
\caption{Ablation study of the architectural components.}
\label{tab:ablation}
\small
\setlength{\tabcolsep}{3pt} 
\resizebox{\columnwidth}{!}{
\begin{tabular}{cccc|cc}
\toprule
\textbf{Base} & \textbf{GCG} & \textbf{MoE} & \textbf{Shared} & \textbf{mAUC} & \textbf{mF1} \\
\midrule
\checkmark & & & & $0.875\pm0.012$ & $0.565\pm0.018$ \\
\checkmark & \checkmark & & & $0.888\pm0.009$ & $0.582\pm0.016$ \\
\checkmark & \checkmark & \checkmark & & $0.901\pm0.009$ & $0.620\pm0.014$ \\
\checkmark & \checkmark & \checkmark & \checkmark & \textbf{0.912$\pm$0.008} & \textbf{0.653$\pm$0.014} \\
\bottomrule
\end{tabular}%
}
\end{table}

\begin{figure*}[t]
  \centering
  \begin{minipage}{0.195\textwidth}\centering
    \includegraphics[width=\linewidth]{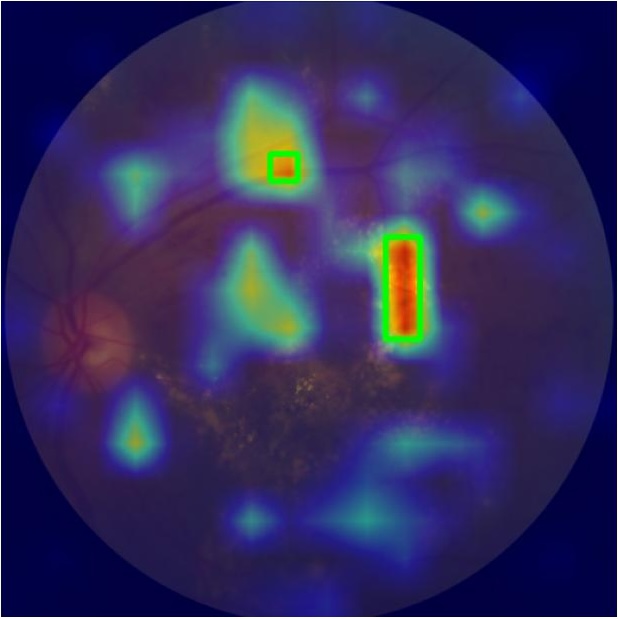}\\[2pt]\small DR
  \end{minipage}\hfill
  \begin{minipage}{0.195\textwidth}\centering
    \includegraphics[width=\linewidth]{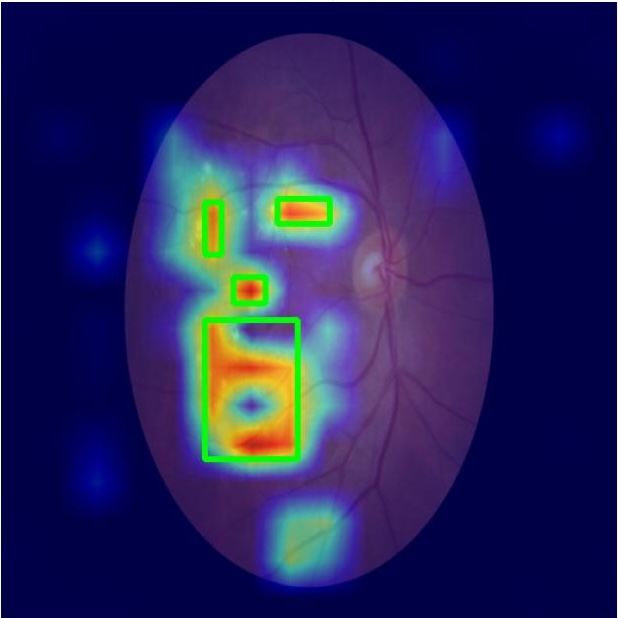}\\[2pt]\small AMD
  \end{minipage}\hfill
  \begin{minipage}{0.195\textwidth}\centering
    \includegraphics[width=\linewidth]{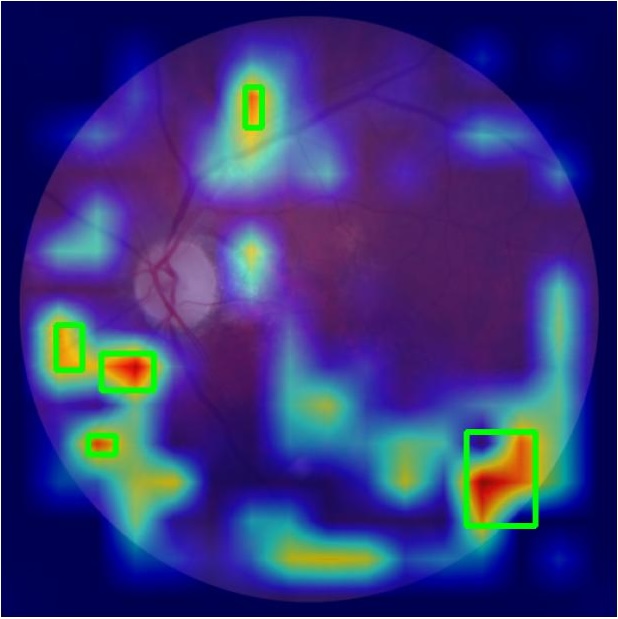}\\[2pt]\small ERM
  \end{minipage}\hfill
  \begin{minipage}{0.195\textwidth}\centering
    \includegraphics[width=\linewidth]{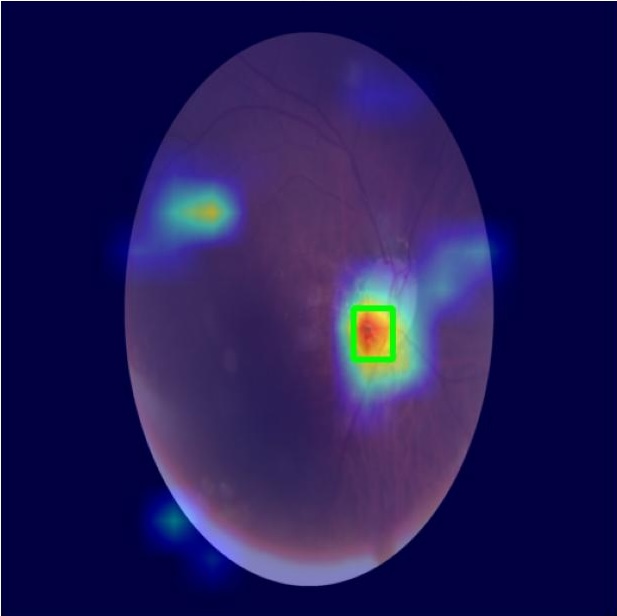}\\[2pt]\small Glaucoma
  \end{minipage}\hfill
  \begin{minipage}{0.195\textwidth}\centering
    \includegraphics[width=\linewidth]{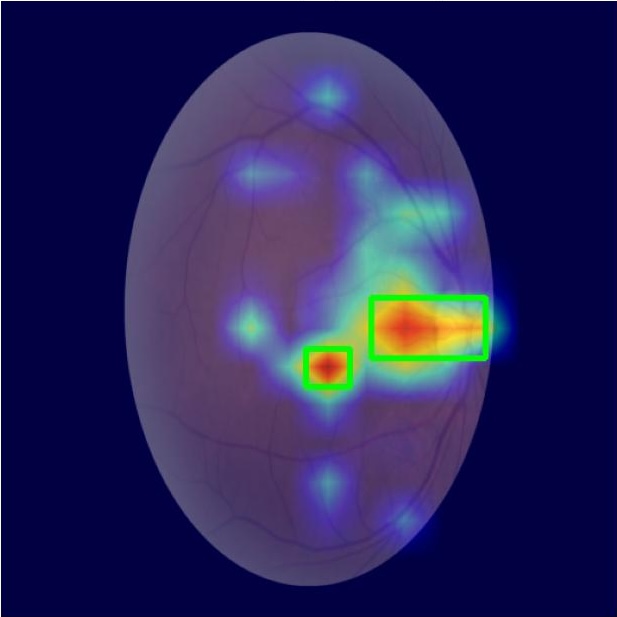}\\[2pt]\small Normal
  \end{minipage}
  \caption{Grad-CAM++ over the gated features. Warmer regions denote higher activation; boxes are salient-region proposals from the thresholded map. DR shows scattered focal activations; AMD and ERM activate the macular/posterior pole; Glaucoma localizes to the optic disc; Normal highlights general anatomy without pathological foci.}
  \vspace{-10pt}
  \label{fig:gradcam}
\end{figure*}

\subsection{Ablation Study}
\label{ssec:ablation}
Table~\ref{tab:ablation} isolates each component added cumulatively to the backbone. GCG provides a strong front-end gain, while the largest F1 improvement comes from the sparse MoE. Adding the shared expert further improves both metrics. The shared expert is a single always-on SwiGLU FFN, increasing active experts per token from top-2 routed to $2$ routed $+$ $1$ shared, thus adding modest capacity rather than keeping the budget fixed. We attribute this gain primarily to routing stabilization: it lowers routing entropy and promotes cleaner per-disease specialization (Sec.~\ref{ssec:qual}) by capturing ubiquitous retinal anatomy, allowing routed experts to focus on disease-specific pathological variation.

\subsection{Qualitative Evaluation}
\label{ssec:qual}

\noindent\textbf{Lesion localization (Grad-CAM++).}
Fig.~\ref{fig:gradcam} shows Grad-CAM++ \cite{chattopadhay2018grad} at the GCG block for representative correctly-classified cases. The gated features attend to clinically relevant structures prior to routing: DR exhibits scattered focal activations corresponding to hemorrhages and exudates; AMD and ERM show strong activation in the macular region; Glaucoma localizes precisely to the optic disc; and the Normal case highlights general structural anatomy without pathological foci.

\begin{figure}[t]
  \centering
  \includegraphics[width=0.49\linewidth]{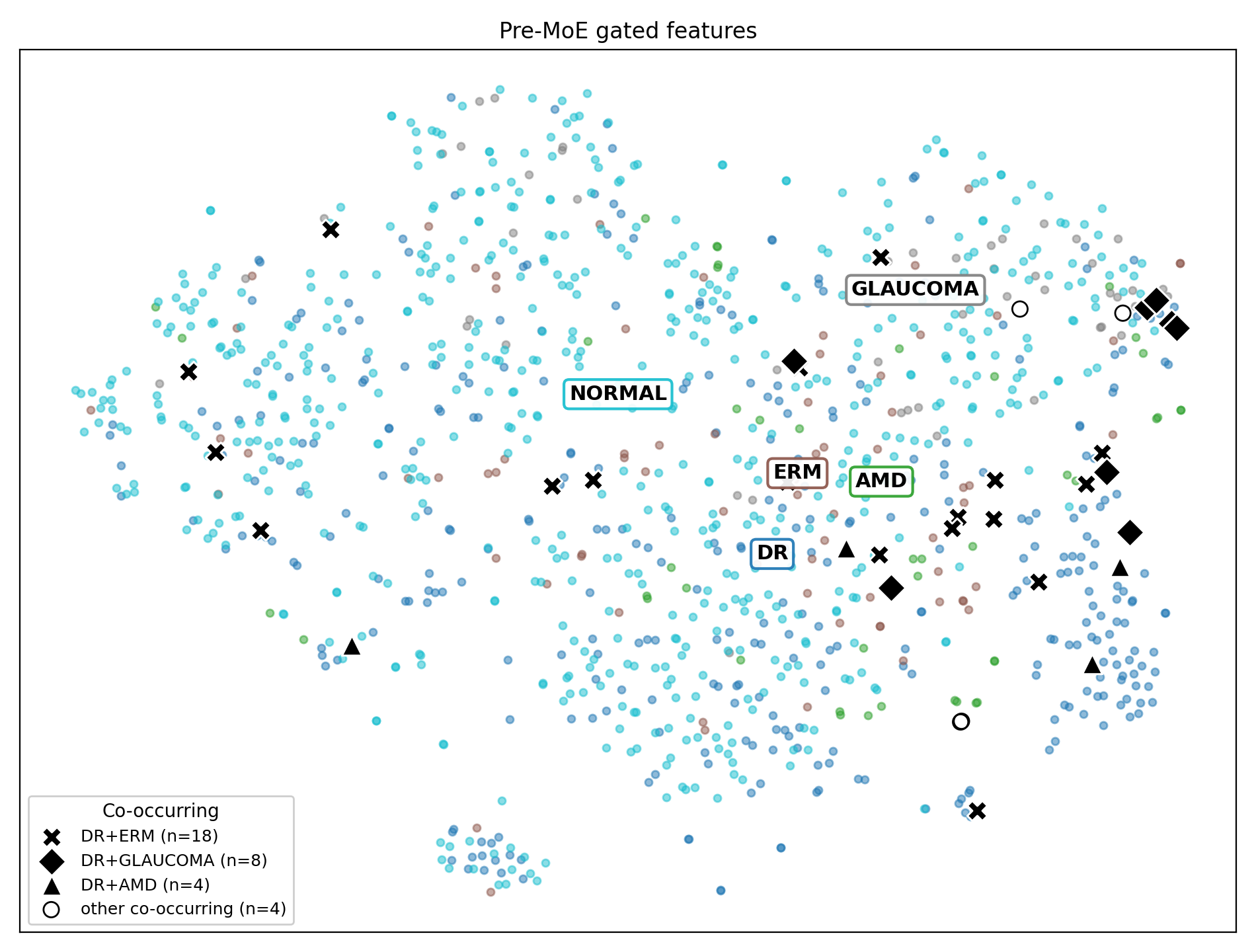}\hfill
  \includegraphics[width=0.49\linewidth]{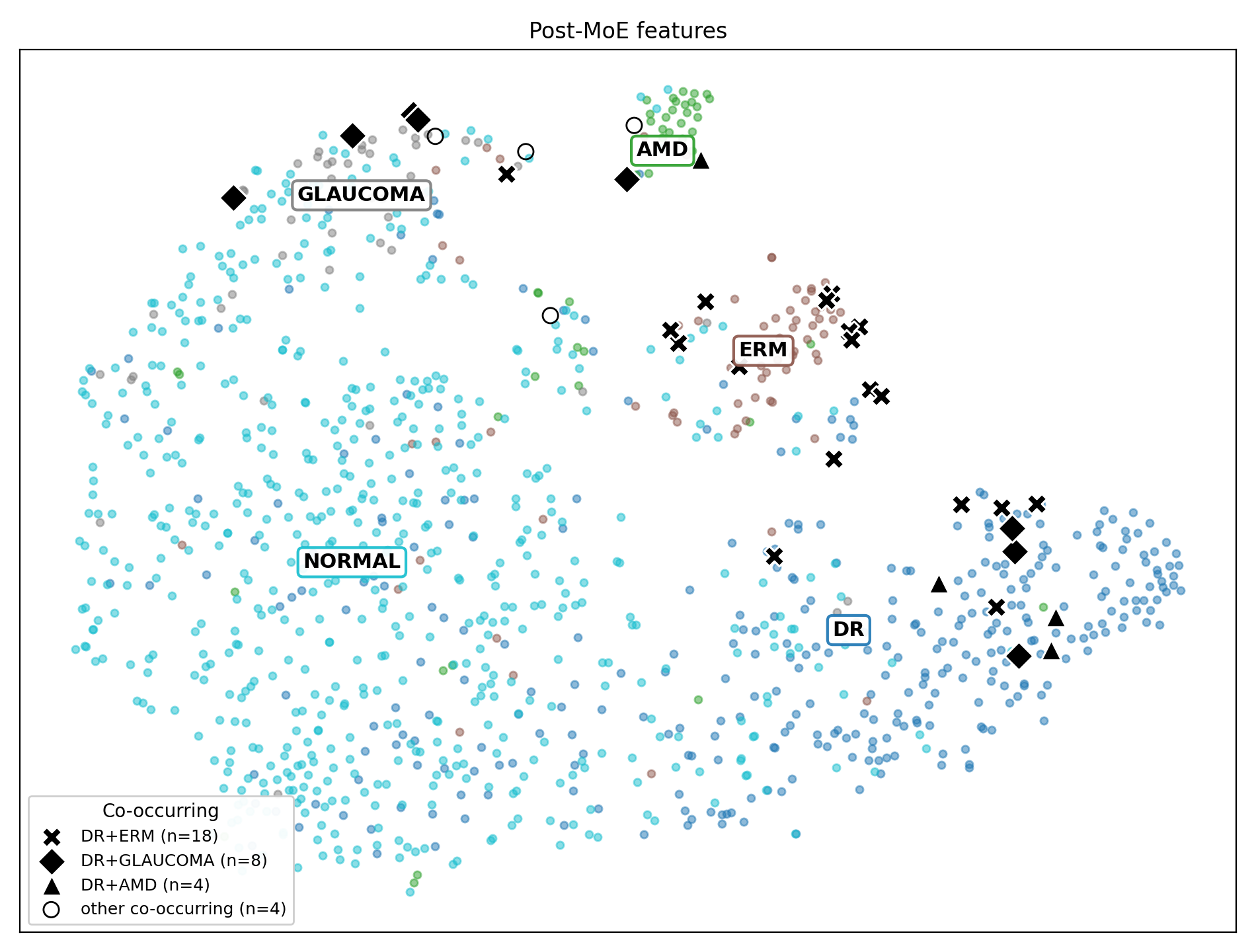}
  \caption{t-SNE of embeddings before (left) and after (right) the MoE block. Single-label diseases are labeled in place; co-occurring cases are drawn as distinct black markers. Post-MoE, AMD and ERM form distinct clusters and co-occurring cases settle between their constituent clusters.}
  \label{fig:tsne}
\end{figure}

\noindent\textbf{Feature disentanglement (t-SNE).}
Fig.~\ref{fig:tsne} compares token-averaged embeddings before and after the MoE block. On single-label test images, the silhouette score over the five classes rises from $-0.098$ (pre-MoE) to $+0.075$ (post-MoE). The negative pre-MoE value is visually evident on the left: the gated features are highly entangled, with all five classes intermixed in one diffuse mass. After routing (right), the space reorganizes: AMD separates into a discernible cluster, ERM consolidates into its own region, Glaucoma separates toward the periphery, and DR resolves into a coherent manifold. Crucially, co-occurring cases (distinct black markers) are not scattered at random but fall in the corridors \emph{between} their constituent single-disease clusters (e.g., DR$+$ERM cases lie along the DR–ERM boundary). The transition from a negative to a positive silhouette is a measurable, data-driven reduction in entanglement.

\noindent\textbf{Expert allocation.}
We test routing disease-dependence on single-label images using mean pairwise Jensen--Shannon divergence (JSD) against a label-permutation null ($2000$ permutations). Routing is significantly disease-dependent (observed $\mathrm{JSD}=0.054$ vs.\ null $0.004$, $p<0.001$). Specialization is \emph{distributed}: morphologically focal pathologies (ERM, Glaucoma, AMD) exhibit low routing entropy and isolate to dominant experts, whereas diffuse states (DR, Normal) route broadly through a shared generalist pathway. This aligns with clinical intuition that focal lesions admit more dedicated representations.

\noindent\textbf{Routing under co-occurrence.}
For dual-label images, we compare joint routing distributions to single-disease prototypes using $1-\mathrm{JSD}$ similarity. For the most frequent pairs, DR$+$ERM ($n=18$) and DR$+$Glaucoma ($n=8$), joint routing strongly matches the \emph{union} of their constituent prototypes (similarities $0.978$ and $0.969$, respectively), exceeding similarity to individual diseases. Thus, co-occurring pathologies do not trigger a generic average, but activate a precise blend of constituent experts, directly proving that sparse routing decomposes multi-label presentations.

\section{Conclusion}
\label{sec:conclusion}
In this work, we introduced a dynamically-routed architecture to address the representation bottleneck in multi-label retinal screening. By fusing a Saliency-Guided Context Gating front-end with a Shared-Expert Sparse MoE block, our model reduces representational disentanglement among co-occurring pathologies. We demonstrated that adapting modern LLM conditional computation techniques to medical vision not only yields state-of-the-art performance on this benchmark but also transforms the routing distribution into an interpretable diagnostic signal. 



\bibliographystyle{IEEEbib}
\bibliography{refs}

\end{document}